\documentclass[
]{ceurart}

\usepackage{listings}
\usepackage{xcolor}

\begin{document}

\copyrightyear{2026}
\copyrightclause{Copyright for this paper by its authors.
  Use permitted under Creative Commons License Attribution 4.0
  International (CC BY 4.0).}

\conference{GenAIECommerce'26: The Third Workshop on Agentic and Generative AI for E-Commerce,
  co-located with RecSys, September 28, 2026, Minneapolis, MN, USA}
\title{What Matters for LLM Persona Simulation: Signal Form in Cold-Start E-Commerce Advertising}


\author[1]{Jiwoo Choi}[
email=choijiwoo@dosan.io
]
\fnmark[1]

\author[1]{In Ik Lee}[
email=dldlsdlr@dosan.io
]
\fnmark[1]

\author[1]{Woo Sung Seo}[
email=tjdntjd777@dosan.io
]
\fnmark[1]

\author[2]{Minho Lee}[
email=mhlee@solverx.ai
]

\author[2]{Yun Young Choi}[
email=young@solverx.ai
]
\cormark[1]

\address[1]{Dosan International, South Korea}
\address[2]{SolverX, South Korea}

\fntext[fn1]{These authors contributed equally.}
\cortext[cor1]{Corresponding author.}

\begin{abstract}
Estimating advertising performance in cold-start settings is challenging because historical logs are unavailable before launch. Large language models (LLMs) offer a potential alternative by simulating persona-conditioned user responses, but it remains unclear which design choices actually drive LLM-based persona simulation. Using real Google Ads e-commerce campaign data across five demographic segments, we identify three signal-form dimensions that affect simulation quality: (i) a demographic anchor without noisy rich attributes, (ii) a relative qualitative banner-blindness signal rather than quantitative anchors, and (iii) prompt phrasing. On this campaign, our method attains an MAE of 24.9 and a MAPE of 26.1\%, a 51.3\% lower MAE than the best tested non-LLM baseline and 77.3\% lower than an LLM configuration using explicit CTR anchors. 
Systematic ablations show that minimal demographic descriptions achieve lower mean error than richer persona configurations, while relative qualitative cues substantially outperform the tested quantitative anchors. We additionally compare these configurations with direct-prior and rule-based baselines and with alternative LLM configurations.
\end{abstract}

\maketitle


\section{Introduction}
\label{sec:introduction}

Online advertising has become a central channel for e-commerce customer acquisition, with advertisers increasingly relying on data-driven audience targeting to allocate campaign budgets~\cite{perlich2014targeted}. However, for new products or campaigns without historical interaction logs, estimating how each audience segment will respond remains a persistent open problem. While conventional click-through rate (CTR) prediction models rely on historical user--ad interactions, such signals may be unavailable or unreliable for new campaigns, products, or creative assets~\cite{geng2024breaking,cui2025field}.

Large language models (LLMs) offer a new opportunity for cold-start audience evaluation. By conditioning an LLM on a target persona and an ad creative, one can simulate whether a synthetic user would click the ad and aggregate repeated simulations into segment-level response estimates. This approach requires no user-level behavioral logs and can incorporate natural-language descriptions of products, creatives, and audience segments. Recent work has explored LLMs for persona-conditioned human behavior simulation and online shopping behavior modeling~\cite{hu2024persona,wang2025customerr1}, as well as LLM-enhanced CTR prediction and advertising retrieval~\cite{geng2024breaking,cui2025field,liu2025rare}.

However, it remains unclear which design choices actually drive LLM-based persona simulation for advertising response estimation. Advertising clicks are not ordinary preference judgments: even when users are interested in a product, display ads are frequently ignored due to banner blindness, attention scarcity, and the low-base-rate nature of ad clicks. To address this, domain knowledge can be injected into simulation prompts, such as banner-blindness cues and low-base-rate CTR priors. However, we find that injecting more domain knowledge is not always better---and \emph{how} the knowledge is expressed can matter more than \emph{how much} is provided.

\begin{figure}[t]
  \centering
  \includegraphics[width=\linewidth]{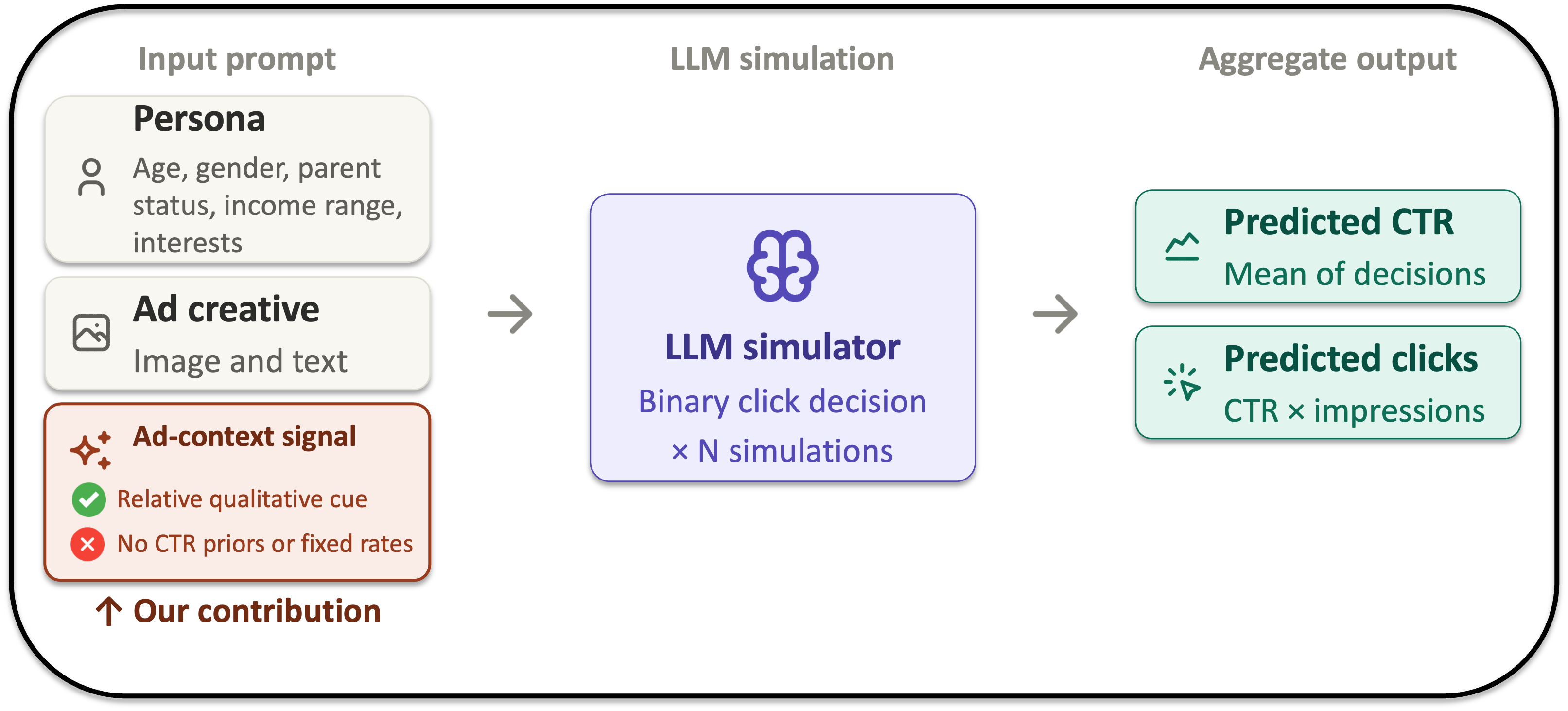}
  \caption{Overview of our LLM persona simulation framework. We examine three key dimensions of signal form that affect simulation quality: (i) a demographic anchor, (ii) a relative qualitative banner-blindness signal, and (iii) careful prompt phrasing. The LLM simulator generates repeated binary click decisions, which are aggregated into predicted CTR and predicted clicks.}
  \label{fig:framework}
\end{figure}

In this paper, we study which signal-form choices make LLM-based persona simulation effective for cold-start e-commerce display advertising response estimation. Using real Google Ads campaign data from five demographic audience segments, we ask: \emph{What form of signal makes LLM persona simulation effective, and how does its accuracy compare to non-LLM baselines on the same data?} Our method combines the base persona with a generic banner-blindness
instruction and a relative qualitative banner-blindness cue, without
quantitative anchors or rich persona attributes. Figure~\ref{fig:framework} illustrates our simulation framework and the three key elements we identify.

On this campaign, our method attains the lowest error among all tested configurations, with an MAE of 24.9 and a MAPE of 26.1\%---a 51.3\% lower MAE than the best tested non-LLM baseline and 77.3\% lower than an LLM configuration using explicit CTR anchors. 
Three patterns characterize this result: (i) the base demographic persona achieves lower mean error than richer persona configurations; (ii) a relative qualitative signal substantially outperforms the tested quantitative anchors; and (iii) prompt phrasing affects accuracy even when the signal content is held fixed.

Our contributions are as follows:
\begin{itemize}
    \item We formulate cold-start e-commerce display advertising response estimation as an LLM persona simulation task and evaluate it using real Google Ads campaign data.
    \item We identify three signal-form dimensions that affect LLM persona simulation quality in our evaluated setting: (i) persona granularity, (ii) relative qualitative versus quantitative behavioral signals, and (iii) prompt phrasing.
    \item We show that, on this campaign, the resulting minimal configuration attains the lowest error among all tested configurations, with an MAE of 24.9 (a 51.3\% lower MAE than the best tested non-LLM baseline), and we discuss the conditions and limitations relevant to transferring these findings beyond the evaluated campaign.
\end{itemize}

\section{Related Work}

\subsection{LLM-Enhanced CTR Prediction and Advertising Retrieval}

CTR prediction is a central task in online advertising and recommender systems, where models estimate the probability of user engagement from historical user--item or user--ad interactions. Recent studies have explored how LLMs can enhance CTR prediction and advertising retrieval pipelines. For example, LLMs have been used to encode long textual user behavior sequences, extract field-level semantic knowledge, and generate commercial intentions for sponsored search advertising~\cite{geng2024breaking,cui2025field,liu2025rare}. These studies show that LLMs can provide useful semantic signals for advertising systems.

Our work differs from this line of research in both task setting and prediction target. Prior LLM-enhanced CTR methods typically assume access to historical interaction logs and focus on supervised impression-level prediction or retrieval. In contrast, we study cold-start e-commerce display advertising, where historical logs for a new product or campaign may be unavailable. Rather than predicting individual click labels, we estimate aggregate segment-level ad responses through repeated LLM persona simulations.

\subsection{LLM Persona Simulation for Human Behavior}

LLMs have also been studied as simulators of persona-conditioned human behavior. Prior work has examined how persona attributes affect LLM simulations and whether LLM agents can reproduce personalized behaviors in domains such as online shopping~\cite{hu2024persona,wang2025customerr1}. These studies suggest that persona-conditioned prompting can capture certain aspects of human preference and decision-making, while also raising questions about robustness, bias, and the reliability of simulated behavior.

Our setting introduces a distinct challenge: advertising clicks are low-probability exposure-level behaviors rather than direct preference statements. A user may be interested in a product but still ignore a display ad due to banner blindness or limited attention. Realistic ad response simulation therefore requires not only persona modeling, but also calibration to the behavioral context of display advertising. Moreover, we go beyond evaluating whether LLM simulations can reproduce behavior at all, and systematically investigate \emph{which design choices} make persona-conditioned prompting effective in this setting.

\subsection{Prompt Design and Domain Knowledge Injection}

Prompt design has emerged as an important axis of LLM system development. Recent studies have proposed system prompt optimization, retrieval-augmented prompt refinement, multi-agent design, and textual feedback optimization to improve LLM reasoning and task performance~\cite{choi2025system,lee2025better,zhou2025multiagent,yuksekgonul2024textgrad}. These methods suggest that LLM outputs can be substantially affected by prompt design, feedback, or agent coordination.

However, these approaches largely focus on general reasoning improvements rather than domain-specific behavioral calibration. In advertising response simulation, the target behavior is not general preference reasoning but low-base-rate ad clicking under display exposure, and injected signals must reflect the peculiarities of ad-exposure context. Rather than proposing a new prompt optimization algorithm, we empirically study \emph{what form of signal} makes domain knowledge injection effective in this setting---distinguishing between demographic anchoring, relative versus quantitative behavioral cues, and prompt phrasing.

\section{Problem Setting}

We consider a cold-start e-commerce display advertising setting in which an advertiser wants to estimate campaign responses before sufficient historical interaction logs are available. In this paper, response estimation refers specifically to segment-level CTR estimation and the derived estimation of click counts. Let $\mathcal{S}$ denote a set of audience segments, where each segment $s \in \mathcal{S}$ is described by demographic attributes such as age range, gender, parent status, and income range. Given an ad creative $a$ and optional temporal context $c_t$, our goal is to estimate the aggregate advertising response of each segment over time.

We focus specifically on estimating segment-level CTR. The number of impressions allocated to each segment is treated as given, since impression volume is typically planned in advance through budget allocation and bidding rather than predicted by the response model. Under this assumption, predicted clicks are obtained by scaling the estimated CTR by the planned impressions, and the central modeling challenge is to estimate CTR accurately when no historical interaction logs are available.

Unlike conventional CTR prediction, which is typically formulated as impression-level supervised learning over historical user--ad interactions~\cite{geng2024breaking,cui2025field}, our task focuses on segment-level aggregate response estimation. Specifically, we use an LLM to simulate repeated click decisions for a persona representing segment $s$, and then aggregate these simulated decisions into predicted CTR and predicted clicks. The prediction is compared against observed campaign outcomes from Google Ads.

Formally, for each segment $s$ and date-time context $t$, we construct a persona prompt $p_s$. Given persona prompt $p_s$, ad creative $a$, and, when applicable, temporal context $c_t$, the LLM simulator produces a binary click decision for the $i$-th simulation trial:
\[
\hat{y}_{s,t,i} = f_{\theta}(p_s, a, c_t),
\]
where $f_{\theta}$ denotes the fixed pretrained LLM used as the simulator,
and $\hat{y}_{s,t,i} \in \{0,1\}$ indicates whether the simulated persona
clicks the ad. We repeat this simulation $N_{s,t}$ times and aggregate the outputs:
\[
\widehat{\mathrm{CTR}}_{s,t} =
\frac{1}{N_{s,t}}\sum_{i=1}^{N_{s,t}}\hat{y}_{s,t,i}.
\]
Given impressions $I_{s,t}$, predicted clicks are:
\[
\hat{C}_{s,t} = I_{s,t} \cdot \widehat{\mathrm{CTR}}_{s,t}.
\]
The predicted clicks $\hat{C}_{s,t}$ are compared with observed clicks $C_{s,t}$ from campaign data.

In our study, the evaluation data consist of real Google Ads display campaign outcomes from five demographic audience segments. Each segment's demographic attributes serve as the persona description for LLM simulation, while its observed CTR provides the ground truth for evaluation. Table~\ref{tab:personas} summarizes the audience segments used in our experiments, along with their observed click-through rates (CTR).

\begin{table}[t]
\caption{Audience segments used as personas and their observed CTR over the 7-day evaluation period. All segments additionally share the same interests setting (``Shoppers''); income range is part of the base persona and is included in the prompt.}
\label{tab:personas}
\begin{tabular}{lllll}
\toprule
Persona & Age & Gender & Parent status & Actual CTR (\%) \\
\midrule
A & 45--54 & Female & Parent & 0.680 \\
B & 55--64 & Male & Parent & 0.607 \\
C & 25--34 & Female & Non-parent & 0.174 \\
D & 25--34 & Male & Non-parent & 0.200 \\
E & 35--44 & Female & Parent & 0.266 \\
\bottomrule
\end{tabular}
\end{table}

\section{Signal Form-Aware Persona Simulation}

A key challenge in LLM-based ad response simulation is that LLMs may interpret ad clicks as direct expressions of user preference, which is inappropriate for display advertising. Users may ignore banner ads even when the product is relevant, and click events occur at a low base rate. Persona-only prompting can therefore overestimate click propensity, especially when persona descriptions encourage the model to infer strong product interest from demographic or lifestyle attributes.

A natural response is to inject domain knowledge into the prompt---for example, banner-blindness cues or low-base-rate CTR priors~\cite{smartinsights2024displayctr}---to calibrate the simulation toward realistic ad-exposure behavior. However, domain knowledge can be expressed in many different forms, and it is unclear which forms are most effective. We introduce three signal-form components that jointly determine the prompt: the persona description, the ad-context signal, and the prompt phrasing.

\subsection{Framework Overview}

We consider a prompting framework in which the LLM simulator receives an input prompt of the form:
\[
x_{s,t} = [p_s;\ a;\ c_t;\ q;\ d],
\]
where $p_s$ is the persona description for segment $s$, $a$ is the ad creative, $c_t$ denotes optional temporal context included only in datetime-aware ablations, $q$ is the click-decision query, and $d$ denotes the ad-context signal. The LLM produces a binary click decision:
\[
\hat{y}_{s,t,i} = f_{\theta}(x_{s,t}).
\]

The persona description $p_s$ and the ad-context signal $d$ define the two components of $x_{s,t}$ through which signal form can be varied. In addition, the specific phrasing used to express $d$ constitutes an implicit third design choice. We describe each in the next subsection.

\subsection{Signal Form Components}
\label{sec:signal_form}

We use \emph{signal form} to refer to three prompt-design dimensions:
the persona description, the representation of the ad-context signal,
and the phrasing used to express that signal.

\paragraph{Persona Description ($p_s$).}
The persona description can range from a minimal profile to richer ones. We define the base persona as the set of demographic attributes that can be configured as native targeting units in Google Ads: age range, gender, parental status, income range, and interests (fixed to ``Shoppers'' across all segments). Beyond this base, we separately consider two types of augmentation: (i) temporal context (datetime and day of week), and (ii) rich attributes that are not available as native targeting options, such as personality and occupation. These augmentations are varied independently in our ablation.

\paragraph{Ad-Context Signal ($d$).}
The ad-context signal encodes banner-blindness knowledge in the prompt. We use a generic banner-blindness instruction (GBI) that reminds the model that most users ignore display ads and should not default to a positive click decision. On top of this generic instruction, we compare different forms of behavioral signals.

Two families of expression are considered. Quantitative anchors provide precise numerical targets, such as an explicit CTR prior
(``ad CTR is typically 0.5\%''), an absolute banner-blindness statement (``users ignore approximately 90\% of ads''), or a segment-specific CTR table. Relative qualitative signals, in contrast, describe behavioral tendencies without hard numerical targets (e.g., ``younger adults tend to show stronger banner blindness than older adults'').

\paragraph{Prompt Phrasing.}
Given a persona description and an ad-context signal, the specific phrasing of the ad-context instruction remains a design choice. Different phrasings can convey the same qualitative content with varying modality (``exhibit'' vs.\ ``tend to show'') or intensity (``significantly more'' vs.\ ``much stronger''). We treat prompt phrasing as an explicit variable rather than a stylistic detail.

We systematically investigate the impact of each of these three components on simulation quality in Section~\ref{sec:results}.

\subsection{Our Method}
Combining these three components, our method uses:
\begin{itemize}
    \item \textbf{Persona description ($p_s$)}: the base persona only (the Google Ads targeting attributes defined in Section~\ref{sec:signal_form}), without temporal context or additional rich attributes.
    \item \textbf{Ad-context signal ($d$)}: the generic banner-blindness instruction (GBI) together with a relative qualitative banner-blindness cue, without quantitative anchors such as explicit CTR priors, absolute banner-blindness statements, or segment-specific CTR tables.
    \item \textbf{Prompt phrasing}: a soft-modality expression of the banner-blindness cue.
\end{itemize}

Table~\ref{tab:prompt_example} shows the abbreviated structure of our prompt, which elicits a binary click decision from the model.

\begin{table}[t]
\caption{Abbreviated prompt structure of our method.}
\label{tab:prompt_example}
\small
\begin{tabular}{p{0.95\linewidth}}
\toprule
\textbf{Component} \\
\midrule

\textbf{Persona ($p_s$):}
You are a user with the following demographic attributes:
age range, gender, parental status, income range, and interests. \\

\textbf{Ad creative ($a$):}
You are shown the following display ad, including text metadata and image. \\

\textbf{Generic banner-blindness instruction (GBI):}
The ``No-Click'' Default: Acknowledge ``Banner Blindness.''
Most users ignore ads. Do not default to a positive action. \\

\textbf{Relative banner-blindness cue:}
Younger adults tend to show much stronger banner blindness than older adults. \\

\textbf{Query ($q$):}
Would this user click the ad? Return a binary click decision. \\

\bottomrule
\end{tabular}
\end{table}

\section{Experimental Setup}
\label{sec:experimental_setup}

\subsection{Dataset}

We evaluate our approach using real Google Ads display campaign data
for a jelly product across the five demographic audience segments
described in Table~\ref{tab:personas}. Each segment corresponds to a separately targeted campaign group, allowing us to obtain segment-level impressions, clicks, and CTR. The main evaluation uses a seven-day aggregated setting over five persona types, covering the period from April 5 to April 11, 2026. Across this evaluation period, the dataset contains 974,423 impressions and 3,578 observed clicks. The evaluation therefore covers 35 persona-day instances per experiment condition.

\subsection{Baselines and Comparison Configurations}

We compare our method against both non-LLM baselines and alternative LLM signal-form configurations. We focus on baselines that operate under the same cold-start information regime as our method. Standard supervised CTR prediction models typically require historical user--ad interaction logs for model fitting or adaptation, which are unavailable for the target campaign in our setting.

\paragraph{Non-LLM Baselines.}
We include two families of non-LLM baselines:
\begin{itemize}
    \item \textbf{Direct prior}: predicts clicks as $\hat{C}_{s,t} = I_{s,t} \cdot p$ for a fixed CTR prior $p \in \{0.2\%, 0.5\%, 1.0\%\}$, without using persona, creative, or temporal context. This baseline reflects a common industry heuristic that scales predicted clicks by a global display advertising CTR benchmark~\cite{smartinsights2024displayctr}.
    \item \textbf{Rule-based}: predicts CTR as $\widehat{\mathrm{CTR}}_s = p_{\mathrm{base}} \cdot m(\mathrm{age}(s))$, where $p_{\mathrm{base}} = 0.5\%$ is the reference display-ad CTR and $m(\cdot)$ is derived from the age-group CTR profile reported by Beel et al.~\cite{beel2013demographics}. Because the source concerns research paper recommendations rather than display advertising, we use the profile only as a cross-domain age-monotonic heuristic. Specifically, $m_g=r_g/\bar{r}$, where $r_g$ is the reported CTR for age group $g$ and $\bar{r}$ is computed using either an unweighted or a weighted average.
\end{itemize}

\paragraph{LLM Baselines.}
We consider three LLM configurations that vary the persona description $p_s$ and the ad-context signal $d$:
\begin{itemize}
    \item \textbf{Persona + GBI}: prompts that use the base demographic and temporal attributes together with the generic banner-blindness instruction (GBI), but without an additional relative qualitative cue or quantitative CTR anchor.
    \item \textbf{Explicit CTR Prior}: a configuration that augments the base persona with datetime, personality, and occupation, and combines a banner-blindness cue with an explicit low-base-rate CTR prior.
    \item \textbf{Rich Signal}: the same rich persona prompt combined with our relative qualitative banner-blindness cue, used to isolate the effect of persona granularity while holding the ad-context signal and prompt phrasing fixed.
\end{itemize}

\paragraph{Alternative Signal-Form Configurations.}
To isolate the contribution of each component in Section~\ref{sec:signal_form}, we also evaluate alternative configurations, including temporal context (datetime), rich persona attributes (personality and occupation), quantitative variants of the banner-blindness signal (absolute ``90\% ignore'' statements, segment-specific CTR tables), and alternative phrasings of the relative banner-blindness cue.

\subsection{Simulation Procedure}

For each persona, day, and prompting condition, we run repeated LLM simulations using \texttt{gpt-5-nano-2025-08-07} at temperature 1.0. Each prompt asks whether the persona would click the display ad, and the model returns a binary click decision. These decisions are aggregated into simulated CTR and converted into predicted clicks using observed impressions. Datetime information is included only in datetime-aware ablations and is not used in our selected configuration.

For each experiment condition, the total number of simulated instances is 974,423, matching total impressions across five personas over seven days. Datetime-aware conditions include a timestamp in \texttt{YYYY-MM-DD HH:mm} format and the day of week. Each ad creative is represented using textual metadata and the actual display ad image; the same creative information is used across all prompting conditions.

For our main LLM configurations, we conduct three independent runs and report the mean and standard deviation. Non-LLM baselines are deterministic given the CTR prior and impressions.

\subsection{Evaluation Metrics}

We evaluate aggregate click estimation using mean absolute error (MAE), mean absolute percentage error (MAPE), and root mean squared error (RMSE). Given observed clicks $C_j$ and predicted clicks $\hat{C}_j$ for each persona-day instance $j$, the metrics are defined as:
\[
\mathrm{MAE} = \frac{1}{M}\sum_{j=1}^{M} |C_j - \hat{C}_j|,
\]
\[
\mathrm{RMSE} = \sqrt{\frac{1}{M}\sum_{j=1}^{M}(C_j - \hat{C}_j)^2},
\]
\[
\mathrm{MAPE} = \frac{100}{M}\sum_{j=1}^{M}\left|\frac{C_j - \hat{C}_j}{C_j}\right|.
\]
where $M$ denotes the total number of persona-day evaluation instances. Since there are no persona-day instances with zero observed clicks in our evaluation data, MAPE is computed without excluding instances or applying smoothing.

\section{Results and Analysis}
\label{sec:results}

\subsection{Overall Performance}

Table~\ref{tab:main_results} reports aggregate click estimation performance across the seven-day evaluation period. Among all tested configurations, our method attains the lowest error on every aggregate metric, with an MAE of $24.9 \pm 1.9$, a MAPE of $26.1\% \pm 0.9$, and an RMSE of $27.0 \pm 1.8$. This corresponds to a 51.3\% lower MAE than the best tested non-LLM baseline (Direct Prior 0.2\%, MAE 51.1) and a 77.3\% lower MAE than the Explicit CTR Prior configuration (MAE 109.9). Across the three independent runs, our method shows modest variability, with standard deviations below 8\% of the corresponding mean values.

\begin{table*}[t]
\caption{Main results on 7-day evaluation (5 personas, 974,423 impressions, 3,578 clicks). Our method attains the lowest error among all tested configurations on this campaign. LLM methods report mean $\pm$ SD over 3 independent runs; non-LLM baselines are deterministic. Best values on aggregate metrics in \textbf{bold}. $\downarrow$: lower is better.}
\label{tab:main_results}
\centering
\small
\begin{tabular}{lrrr}
\toprule
Method & MAE $\downarrow$ & MAPE (\%) $\downarrow$ & RMSE $\downarrow$ \\
\midrule
\multicolumn{4}{l}{\textit{Non-LLM Baselines}} \\
Direct Prior 0.2\% & 51.1 & 37.1 & 73.0 \\
Direct Prior 0.5\% & 66.4 & 93.5 & 75.5 \\
Direct Prior 1.0\% & 176.2 & 252.0 & 198.0 \\
Rule-based (avg) & 54.5 & 81.7 & 63.8 \\
Rule-based (weighted avg) & 57.5 & 85.9 & 58.3 \\
\midrule
\multicolumn{4}{l}{\textit{LLM Baselines} (mean $\pm$ SD, 3 runs)} \\
Persona + GBI & 153.5 $\pm$ 16.6 & 212.9 $\pm$ 32.4 & 155.4 $\pm$ 16.9 \\
Explicit CTR Prior & 109.9 $\pm$ 1.6 & 127.7 $\pm$ 3.2 & 112.0 $\pm$ 1.2 \\
Rich Signal & 37.7 $\pm$ 3.0 & 38.0 $\pm$ 2.4 & 39.3 $\pm$ 3.0 \\
\midrule
\multicolumn{4}{l}{\textit{Ours} (mean $\pm$ SD, 3 runs)} \\
\textbf{Ours} & \textbf{24.9 $\pm$ 1.9} & \textbf{26.1 $\pm$ 0.9} & \textbf{27.0 $\pm$ 1.8} \\
\bottomrule
\end{tabular}
\end{table*}

Three patterns stand out from these results. First, the Explicit CTR Prior configuration is substantially worse than simple non-LLM baselines on aggregate accuracy, despite being provided with a precise numerical CTR target. Second, the Rich Signal configuration reduces MAE from 109.9 to 37.7 by replacing the quantitative CTR anchor with a relative qualitative banner-blindness signal, while keeping the persona configuration fixed. Third, reducing the persona to the base persona alone attains the lowest error across all three aggregate metrics among the tested configurations. Together, these patterns describe a consistent ordering of signal forms, which we analyze element by element in the following subsections.

\subsection{Effect of Removing the Generic Banner-Blindness Instruction}

To isolate the effect of the generic banner-blindness
instruction, we conduct an additional ablation in which this instruction is removed while the model, persona, and simulation configurations are otherwise unchanged. This setting is distinct from the corresponding persona baseline in Table~\ref{tab:main_results}, which retains the generic banner-blindness instruction.

As shown in Table~\ref{tab:no_banner_ratio}, removing the generic
banner-blindness instruction causes severe click inflation. Even the least inflated rich persona + datetime condition predicts 910,634 clicks against 3,578 observed clicks, while the demographics-only condition predicts 947,516 clicks.

\begin{table}[t]
\caption{Click inflation after removing the generic banner-blindness instruction. The model, persona, and simulation
configurations are otherwise unchanged. Ratio $= \hat{C}/C$.}
\label{tab:no_banner_ratio}
\centering
\small
\begin{tabular}{lrrr}
\toprule
Prompt & Actual clicks & Pred. clicks & Ratio \\
\midrule
Demographics & 3,578 & 947,516 & 264.82 \\
Demographics + datetime & 3,578 & 937,482 & 262.01 \\
Rich persona & 3,578 & 924,119 & 258.28 \\
Rich persona + datetime & 3,578 & 910,634 & 254.51 \\
\bottomrule
\end{tabular}
\end{table}

These results confirm that removing the generic banner-blindness
instruction causes the LLM to treat ad engagement as a high-probability preference event rather than a rare exposure-level behavior. This motivates all subsequent signal-form analyses, which assume that banner blindness in some form is included and instead ask \emph{how} the signal should be expressed.

\subsection{Signal Form Analysis}

We now examine each of the three signal-form components introduced in Section~\ref{sec:signal_form}. Table~\ref{tab:signal_form} reports a unified ablation over the three components: persona granularity (rows), prompt phrasing (columns), and signal type (the two blocks). Each configuration is evaluated over three independent runs, with results reported as mean and standard deviation.

\subsubsection{Element 1: Demographic Anchor}
The persona rows of Table~\ref{tab:signal_form} isolate the effect of
persona granularity, with the signal type and phrasing held fixed.
On average, adding persona attributes increases error under both
phrasings. Under v2, MAE increases from $24.9 \pm 1.9$ for the base
persona to $26.1 \pm 3.4$ with datetime and personality, and further
to $37.7 \pm 3.0$ when occupation is added. A similar pattern is
observed under v1. These results suggest that additional persona
attributes do not necessarily improve calibration, with the clearest
degradation occurring for the richest persona configuration.

\subsubsection{Element 2: Relative vs.\ Quantitative Signal}

The two blocks of Table~\ref{tab:signal_form} isolate the effect of signal type while holding the rich persona configuration fixed. The relative qualitative banner-blindness cue achieves an MAE of $37.7 \pm 3.0$, whereas the explicit quantitative CTR anchor yields $109.9 \pm 1.6$---despite providing the model with a more specific numerical target. We observed the same pattern for other quantitative forms in a one-day setting, including general CTR knowledge, absolute banner-blindness statements (``users ignore approximately 90\% of ads''), and segment-specific CTR tables, all of which produced substantially higher error than the relative signal. Precise numerical anchors do not translate into better calibration; instead, they appear to distort the model's default click propensity in ways that substantially degrade aggregate accuracy, as reflected in the high error of the Explicit CTR Prior configuration in Table~\ref{tab:main_results}.

\subsubsection{Element 3: Prompt Phrasing}

The v1 and v2 columns of Table~\ref{tab:signal_form} isolate the effect of prompt phrasing, with the persona and signal type held fixed. We compare two phrasings of the relative banner-blindness cue:
\begin{itemize}
    \item \textbf{v1}: ``Younger adults exhibit significantly more banner blindness than older adults.''
    \item \textbf{v2}: ``Younger adults tend to show much stronger banner blindness than older adults.''
\end{itemize}
Both convey the same qualitative direction but differ in modality (``exhibit'' vs.\ ``tend to show'') and intensity (``significantly more'' vs.\ ``much stronger''). Across persona configurations, the mean MAE differs by approximately 8--30\% between the two phrasings, even though they convey identical qualitative content. We interpret this less as evidence that one phrasing is inherently superior, and more as a cautionary finding: prompt-based advertising simulation can be sensitive to wording alone, which poses a reproducibility risk for methods that rely on a single hand-chosen phrasing. Every phrasing variant still achieves lower MAE than both baselines, so the advantage of the approach itself does not hinge on the choice of phrasing. The effect nonetheless suggests that phrasing should be treated as an explicit design choice rather than a stylistic detail.

\begin{table}[t]
\caption{Ablation over the three signal-form components on the 7-day
evaluation. Results report mean $\pm$ SD over three independent runs.
Persona granularity varies across rows and signal type across the two
blocks, while the two columns report the v1 and v2 phrasings of the
banner-blindness cue.}
\label{tab:signal_form}
\centering
\small
\begin{tabular}{lrr}
\toprule
Persona configuration \& signal & v1 MAE & v2 MAE \\
\midrule
\multicolumn{3}{l}{\textit{Relative qualitative banner-blindness}} \\
\quad Base persona only
    & $35.2 \pm 5.9$ & $\mathbf{24.9 \pm 1.9}$ \\
\quad Base + datetime + personality
    & $37.4 \pm 1.3$ & $26.1 \pm 3.4$ \\
\quad Base + datetime + personality + occupation
    & $41.1 \pm 2.9$ & $37.7 \pm 3.0$ \\
\midrule
\multicolumn{3}{l}{\textit{Explicit CTR prior (quantitative)}} \\
\quad Base + datetime + personality + occupation
    & \multicolumn{2}{c}{$109.9 \pm 1.6$} \\
\bottomrule
\end{tabular}
\end{table}

\subsection{Persona-Level Calibration}

To understand how our method affects segment-level calibration, we compare per-persona predicted CTR against observed CTR for our method and the Explicit CTR Prior configuration. Table~\ref{tab:persona_ctr} reports the results.

\begin{table}[t]
\caption{Per-persona CTR calibration for our method and the Explicit CTR Prior configuration. Predicted and actual CTR are given as percentages, and the ratio $\widehat{\mathrm{CTR}} / \mathrm{CTR}$ measures over- or under-estimation, with values closer to $1.0$ indicating better calibration.}
\label{tab:persona_ctr}
\centering
\small
\begin{tabular}{lrrrrr}
\toprule
 & & \multicolumn{2}{c}{Explicit CTR Prior} & \multicolumn{2}{c}{Ours} \\
\cmidrule(lr){3-4} \cmidrule(lr){5-6}
Persona & Actual CTR & Pred. & Ratio & Pred. & Ratio \\
\midrule
A (45--54 F Parent) & 0.680 & 0.128 & 0.19 & 0.497 & 0.73 \\
B (55--64 M Parent) & 0.607 & 0.062 & 0.10 & 0.703 & 1.16 \\
C (25--34 F Non-parent) & 0.174 & 0.447 & 2.56 & 0.260 & 1.50 \\
D (25--34 M Non-parent) & 0.200 & 0.313 & 1.56 & 0.171 & 0.85 \\
E (35--44 F Parent) & 0.266 & 0.194 & 0.73 & 0.223 & 0.84 \\
\bottomrule
\end{tabular}
\end{table}

Our method improves calibration on every persona relative to the Explicit CTR Prior configuration. In particular, personas A and B---which are severely under-estimated by the Explicit CTR Prior configuration, with ratios of 0.19 and 0.10---move much closer to the observed CTR under our method, reaching ratios of 0.73 and 1.16. Some segment-level errors remain: persona A continues to be under-estimated, with a ratio of 0.73, indicating that the highest-CTR older segment is still difficult to fully calibrate. Overall, however, the persona-level results show that the improvements on aggregate MAE in Table~\ref{tab:main_results} are not driven by aggregate averaging alone, but reflect genuine calibration gains on individual segments.

\subsection{CTR Prior Sensitivity}

Finally, we examine how sensitive aggregate accuracy is to the assumed CTR prior. Table~\ref{tab:prior_sensitivity} varies the CTR prior for both the direct-prior baseline and the Explicit CTR Prior configuration.

\begin{table}[t]
\caption{Sensitivity to CTR prior. Direct prior predicts clicks as impressions multiplied by a fixed CTR. Lower is better.}
\label{tab:prior_sensitivity}
\centering
\small
\begin{tabular}{llrrr}
\toprule
Method & CTR prior & MAE & MAPE (\%) & RMSE \\
\midrule
Direct prior & 0.2\% & \textbf{51.1} & \textbf{37.1} & \textbf{73.0} \\
Direct prior & 0.5\% & 66.4 & 93.5 & 75.5 \\
Direct prior & 1.0\% & 176.2 & 252.0 & 198.0 \\
\midrule
Explicit CTR Prior & 0.2\% & 82.3 & 76.0 & 99.4 \\
Explicit CTR Prior & 0.5\% & 109.9 & 127.7 & 112.0 \\
Explicit CTR Prior & 1.0\% & 129.2 & 169.3 & 153.4 \\
\bottomrule
\end{tabular}
\end{table}

Direct priors are strong aggregate baselines when the assumed CTR is close to the realized campaign base rate: the 0.2\% direct prior achieves MAE 51.1, which happens to align well with the lower-CTR personas in our data. However, this result should not be interpreted as a readily deployable cold-start solution, because in practice advertisers do not know in advance whether the appropriate prior is 0.2\%, 0.5\%, or 1.0\%. The steep degradation between 0.5\% (MAE 66.4) and 1.0\% (MAE 176.2) highlights prior selection as a central challenge for direct-prior methods.

The Explicit CTR Prior configuration shows a similar sensitivity pattern but at a substantially worse operating point: even with the best-tuned 0.2\% prior, it achieves only MAE 82.3, still worse than the direct-prior baseline and far above our method. This indicates that injecting a quantitative CTR anchor into an LLM prompt does not overcome the underlying prior-selection challenge; it simply moves that challenge into the prompt.

\section{Discussion and Limitations}
\label{sec:discussion}

\subsection{Why Signal Form Matters}

Our results suggest that the effectiveness of LLM persona simulation depends less on how much domain knowledge is injected and more on how it is expressed. Richer persona configurations tend to degrade calibration, likely because they encourage the model to over-interpret segment-level preferences in conflict with the low-base-rate nature of ad clicks. Quantitative anchors---including explicit CTR priors, absolute banner-blindness rates, and segment-specific CTR tables---also underperform relative qualitative cues, as the model may interpret them as fixed prediction targets rather than contextual guidance. Relative cues, by contrast, preserve segment differentiation without imposing a specific numerical target.

Small changes in prompt phrasing shift mean MAE by approximately 8--30\% despite preserving the same qualitative meaning. This sensitivity suggests that prompt phrasing should be considered explicitly rather than treated as a purely stylistic detail. Taken together, our results indicate that a minimal demographic anchor combined with a relative qualitative cue may provide a useful starting point, although its effectiveness should be validated across campaigns and models.

\subsection{Comparison with Non-LLM Baselines}

Non-LLM baselines are stronger than they may first appear, but their strengths are conditional. Direct-prior baselines are competitive only when the assumed CTR happens to match the realized campaign base rate---a value that is unavailable in a genuine cold-start setting. Rule-based baselines perform reasonably in our campaign because the observed audience response is broadly age-monotonic and the applied age--CTR profile encodes a similar ordering. This assumption, however, may not hold across products, creatives, or audience compositions.

On this campaign, our method achieves an MAE of 24.9, compared with 51.1 for the best direct-prior baseline and 54.5 for the rule-based baseline. Because the rule-based method already captures the broad age-based ordering, the remaining performance gap primarily reflects improved calibration of absolute click propensity rather than recovery of the segment ranking alone. Importantly, our method does not require either a pre-specified CTR value or an externally specified age-based CTR profile.

\subsection{Limitations}

This study focuses on a controlled analysis of signal-form choices within a single Google Ads e-commerce campaign comprising five demographic segments. This setting enables a detailed comparison across persona configurations, signal types, and prompt phrasings, but further evaluation is needed to determine how the observed patterns transfer across product categories, creative formats, audience compositions, and LLMs.

Because the configurations were compared using the same seven-day campaign data, the reported results are best interpreted as an in-campaign analysis of the relative behavior of different signal forms rather than as an estimate of out-of-campaign performance. Nevertheless, the consistent ordering observed across multiple ablations provides initial evidence that minimal demographic descriptions and relative qualitative cues may be useful design choices. Future work should evaluate this ordering on held-out campaigns with signal configurations fixed in advance, including campaigns whose audience response is not age-monotonic.

The observed 8--30\% variation in mean MAE across prompt phrasings further highlights the importance of reporting and systematically evaluating prompt wording. Finally, the remaining under-estimation for the highest-CTR persona suggests an opportunity to improve calibration for high-response audience segments.

\section{Conclusion}

We studied LLM-based persona simulation for cold-start e-commerce display advertising response estimation using real Google Ads campaign data across five demographic audience segments. Rather than proposing a new prompt optimization algorithm or a heavier calibration framework, we focused on a more basic question: which form of signal makes LLM persona simulation effective.

Our experiments identify three signal-form dimensions that affect simulation quality in the evaluated campaign: (i) a demographic anchor without noisy rich attributes, (ii) a relative qualitative banner-blindness signal rather than quantitative anchors, and (iii) careful prompt phrasing. On our campaign, a minimal configuration combining these three elements attains the lowest error among all tested configurations, with an MAE of 24.9 and a MAPE of 26.1\% (a 51.3\% lower MAE than the best tested non-LLM baseline and 77.3\% lower than an LLM configuration using explicit CTR anchors).

Beyond the specific numbers, our results argue for a shift in how LLM persona simulations for advertising are designed and evaluated. 
Injecting more domain knowledge, or richer persona descriptions, is not automatically better; in our setting, both can degrade performance. Instead, careful attention to the \emph{form} of the injected signal---demographic granularity, qualitative versus quantitative expression, and prompt phrasing---appears to be a more productive lever.

\bibliographystyle{ceurart-num}
\bibliography{sample-ceur}

\begin{thebibliography}{12}
\expandafter\ifx\csname natexlab\endcsname\relax\def\natexlab#1{#1}\fi
\providecommand{\url}[1]{\texttt{#1}}
\providecommand{\href}[2]{#2}
\providecommand{\path}[1]{#1}
\providecommand{\DOIprefix}{doi:}
\providecommand{\ArXivprefix}{arXiv:}
\providecommand{\URLprefix}{URL: }
\providecommand{\Pubmedprefix}{pmid:}
\providecommand{\doi}[1]{\href{http://dx.doi.org/#1}{\path{#1}}}
\providecommand{\Pubmed}[1]{\href{pmid:#1}{\path{#1}}}
\providecommand{\bibinfo}[2]{#2}
\ifx\xfnm\relax \def\xfnm[#1]{\unskip,\space#1}\fi
\bibitem[{Perlich et~al.(2014)Perlich, Dalessandro, Raeder, Stitelman, and
  Provost}]{perlich2014targeted}
\bibinfo{author}{C.~Perlich}, \bibinfo{author}{B.~Dalessandro},
  \bibinfo{author}{T.~Raeder}, \bibinfo{author}{O.~Stitelman},
  \bibinfo{author}{F.~Provost},
\newblock \bibinfo{title}{Machine learning for targeted display advertising:
  Transfer learning in action},
\newblock \bibinfo{journal}{Machine Learning} \bibinfo{volume}{95}
  (\bibinfo{year}{2014}) \bibinfo{pages}{103--127}.
\bibitem[{Geng et~al.(2024)Geng, Huan, Zhang, He, Zhang, Yuan, Zhou, and
  Mo}]{geng2024breaking}
\bibinfo{author}{B.~Geng}, \bibinfo{author}{Z.~Huan},
  \bibinfo{author}{X.~Zhang}, \bibinfo{author}{Y.~He},
  \bibinfo{author}{L.~Zhang}, \bibinfo{author}{F.~Yuan},
  \bibinfo{author}{J.~Zhou}, \bibinfo{author}{L.~Mo},
\newblock \bibinfo{title}{Breaking the length barrier: {LLM}-enhanced {CTR}
  prediction in long textual user behaviors},
\newblock in: \bibinfo{booktitle}{Proceedings of the 47th International ACM
  SIGIR Conference on Research and Development in Information Retrieval},
  \bibinfo{publisher}{Association for Computing Machinery},
  \bibinfo{address}{New York, NY, USA}, \bibinfo{year}{2024}. \URLprefix
  \url{https://doi.org/10.1145/3626772.3657974}.
  \DOIprefix\doi{10.1145/3626772.3657974}.
\bibitem[{Cui et~al.(2025)Cui, Liu, Chen, Lou, Zhang, Wang, Sun, Yang, and
  Wang}]{cui2025field}
\bibinfo{author}{Y.~Cui}, \bibinfo{author}{F.~Liu}, \bibinfo{author}{J.~Chen},
  \bibinfo{author}{X.~Lou}, \bibinfo{author}{C.~Zhang},
  \bibinfo{author}{J.~Wang}, \bibinfo{author}{Y.~Sun},
  \bibinfo{author}{X.~Yang}, \bibinfo{author}{C.~Wang}, \bibinfo{title}{Field
  matters: A lightweight {LLM}-enhanced method for {CTR} prediction},
  \bibinfo{year}{2025}. \URLprefix \url{https://arxiv.org/abs/2505.14057}.
  \href{http://arxiv.org/abs/2505.14057}{{\tt arXiv:2505.14057}}.
\bibitem[{Hu and Collier(2024)}]{hu2024persona}
\bibinfo{author}{T.~Hu}, \bibinfo{author}{N.~Collier},
  \bibinfo{title}{Quantifying the persona effect in {LLM} simulations},
  \bibinfo{year}{2024}. \URLprefix \url{https://arxiv.org/abs/2402.10811}.
  \href{http://arxiv.org/abs/2402.10811}{{\tt arXiv:2402.10811}}.
\bibitem[{Wang et~al.(2025)Wang, Lu, Zhang, Huang, and
  Wang}]{wang2025customerr1}
\bibinfo{author}{Z.~Wang}, \bibinfo{author}{Y.~Lu}, \bibinfo{author}{Y.~Zhang},
  \bibinfo{author}{J.~Huang}, \bibinfo{author}{D.~Wang},
  \bibinfo{title}{Customer-{R1}: Personalized simulation of human behaviors via
  {RL}-based {LLM} agent in online shopping}, \bibinfo{year}{2025}. \URLprefix
  \url{https://arxiv.org/abs/2510.07230}.
  \href{http://arxiv.org/abs/2510.07230}{{\tt arXiv:2510.07230}}.
\bibitem[{Liu et~al.(2025)Liu, Wang, Qin, Lu, Chen, Yang, and
  Shu}]{liu2025rare}
\bibinfo{author}{T.~Liu}, \bibinfo{author}{Z.~Wang}, \bibinfo{author}{M.~Qin},
  \bibinfo{author}{Z.~Lu}, \bibinfo{author}{X.~Chen},
  \bibinfo{author}{Y.~Yang}, \bibinfo{author}{P.~Shu},
  \bibinfo{title}{Real-time ad retrieval via {LLM}-generative commercial
  intention for sponsored search advertising}, \bibinfo{year}{2025}. \URLprefix
  \url{https://arxiv.org/abs/2504.01304}.
  \href{http://arxiv.org/abs/2504.01304}{{\tt arXiv:2504.01304}}.
\bibitem[{Choi et~al.(2025)Choi, Baek, and Hwang}]{choi2025system}
\bibinfo{author}{Y.~Choi}, \bibinfo{author}{J.~Baek}, \bibinfo{author}{S.~J.
  Hwang}, \bibinfo{title}{System prompt optimization with meta-learning},
  \bibinfo{year}{2025}. \URLprefix \url{https://arxiv.org/abs/2505.09666}.
  \href{http://arxiv.org/abs/2505.09666}{{\tt arXiv:2505.09666}}.
\bibitem[{Lee et~al.(2025)Lee, Seo, An, Lee, and Bu}]{lee2025better}
\bibinfo{author}{J.~Lee}, \bibinfo{author}{W.~Seo}, \bibinfo{author}{H.~An},
  \bibinfo{author}{S.~Lee}, \bibinfo{author}{Y.~Bu}, \bibinfo{title}{Better by
  comparison: Retrieval-augmented contrastive reasoning for automatic prompt
  optimization}, \bibinfo{year}{2025}. \URLprefix
  \url{https://arxiv.org/abs/2509.02093}.
  \href{http://arxiv.org/abs/2509.02093}{{\tt arXiv:2509.02093}}.
\bibitem[{Zhou et~al.(2025)Zhou, Wan, Sun, Palangi, Iqbal, Vuli{\'c}, Korhonen,
  and Ar{\i}k}]{zhou2025multiagent}
\bibinfo{author}{H.~Zhou}, \bibinfo{author}{X.~Wan}, \bibinfo{author}{R.~Sun},
  \bibinfo{author}{H.~Palangi}, \bibinfo{author}{S.~Iqbal},
  \bibinfo{author}{I.~Vuli{\'c}}, \bibinfo{author}{A.~Korhonen},
  \bibinfo{author}{S.~{\"O}. Ar{\i}k}, \bibinfo{title}{Multi-agent design:
  Optimizing agents with better prompts and topologies}, \bibinfo{year}{2025}.
  \URLprefix \url{https://arxiv.org/abs/2502.02533}.
  \href{http://arxiv.org/abs/2502.02533}{{\tt arXiv:2502.02533}}.
\bibitem[{Yuksekgonul et~al.(2024)Yuksekgonul, Bianchi, Boen, Liu, Huang,
  Guestrin, and Zou}]{yuksekgonul2024textgrad}
\bibinfo{author}{M.~Yuksekgonul}, \bibinfo{author}{F.~Bianchi},
  \bibinfo{author}{J.~Boen}, \bibinfo{author}{S.~Liu},
  \bibinfo{author}{Z.~Huang}, \bibinfo{author}{C.~Guestrin},
  \bibinfo{author}{J.~Zou}, \bibinfo{title}{Textgrad: Automatic
  ``differentiation'' via text}, \bibinfo{year}{2024}. \URLprefix
  \url{https://arxiv.org/abs/2406.07496}.
  \href{http://arxiv.org/abs/2406.07496}{{\tt arXiv:2406.07496}}.
\bibitem[{{Smart Insights}(2024)}]{smartinsights2024displayctr}
\bibinfo{author}{{Smart Insights}}, \bibinfo{title}{Average clickthrough rates
  for paid search, display and social ads},
  \bibinfo{howpublished}{\url{https://www.smartinsights.com/internet-advertising/internet-advertising-analytics/display-advertising-clickthrough-rates/}},
  \bibinfo{year}{2024}. \bibinfo{note}{Accessed: 2026-05-19}.
\bibitem[{Beel et~al.(2013)Beel, Langer, N{\"u}rnberger, and
  Genzmehr}]{beel2013demographics}
\bibinfo{author}{J.~Beel}, \bibinfo{author}{S.~Langer},
  \bibinfo{author}{A.~N{\"u}rnberger}, \bibinfo{author}{M.~Genzmehr},
\newblock \bibinfo{title}{The impact of demographics (age and gender) and other
  user-characteristics on evaluating recommender systems},
\newblock in: \bibinfo{booktitle}{Research and Advanced Technology for Digital
  Libraries}, volume \bibinfo{volume}{8092} of \textit{\bibinfo{series}{Lecture
  Notes in Computer Science}}, \bibinfo{publisher}{Springer},
  \bibinfo{year}{2013}, pp. \bibinfo{pages}{396--400}.
  \DOIprefix\doi{10.1007/978-3-642-40501-3_45}.

\end{thebibliography}

\end{document}